\documentclass[runningheads]{llncs}
\usepackage[T1]{fontenc}
\usepackage{graphicx}
\usepackage{booktabs}
\usepackage{multirow}
\usepackage{float}

\begin{document}
\title{MariNER: A Dataset for Historical Brazilian Portuguese Named Entity Recognition}
\author{João Lucas Luz Lima Sarcinelli\orcidID{0009-0001-4296-0937} \and
Marina Lages Gonçalves Teixeira\orcidID{0000-0003-0471-7714} \and
Jade Bortot de Paiva\orcidID{0009-0007-4962-9415} \and
Diego Furtado Silva\orcidID{0000-0002-5184-9413}}
% %
\authorrunning{J. L. L. L. Sarcinelli et al.}
\institute{USP, São Carlos SP, Brazil
\email{\{joao.luz,marinalages,jade\_bortot\_paiva,diegofsilva\}@usp.br}}
%
% \author{Omitted\inst{1}\orcidID{null}}
% \institute{USP \\ \email{\{joao.luz,marinalages,jade_bortot_paiva,diegofsilva\}@usp.br}}

\maketitle              % typeset the header of the contribution
\begin{abstract}
Named Entity Recognition (NER) is a fundamental Natural Language Processing (NLP) task that aims to identify and classify entity mentions in texts across different categories. While languages such as English possess a large number of high-quality resources for this task, Brazilian Portuguese still lacks in quantity of gold-standard NER datasets, especially when considering specific domains. Particularly, this paper considers the importance of NER for analyzing historical texts in the context of digital humanities. To address this gap, this work outlines the construction of MariNER: \textit{Mapeamento e Anotações de Registros hIstóricos para NER} (Mapping and Annotation of Historical Records for NER), the first gold-standard dataset for early 20th-century Brazilian Portuguese, with more than 9,000 manually annotated sentences. We also assess and compare the performance of state-of-the-art NER models for the dataset.

\keywords{NER \and Historical \and Portuguese \and NLP}
\end{abstract}
\section{Introduction}

Named Entity Recognition (NER) is the task of identifying and classifying mentions of entities within a text into predefined categories, such as persons, locations, organizations, and dates \cite{li_survey_2022}. As a fundamental component of Natural Language Processing (NLP), NER is considered an important step to several downstream applications. Its importance extends to specialized domains, such as medical, legal, digital humanities \cite{tamper_visualizing_2023} domains \cite{oliveira_semclinbr_2022,villavicencio_lener-br_2018,tamper_visualizing_2023}, where the ability to extract and analyse named entities can provide valuable insights from large volumes of text. For instance, automated NER allows large-scale analysis of unstructured historical texts in digital humanities. It enables researchers to analyze trends, map historical events, and uncover relevant patterns \cite{ehrmann_named_2024}. 

However, developing effective NER systems relies heavily on the availability of high-quality, annotated datasets to train entity extraction models. For many languages, including Portuguese, such resources are scarce for some domain-specific contexts, such as historical texts \cite{albuquerque_named_2023}, where the linguistic and contextual nuances of the data, as well as availability and the necessity to digitize documents, may require specialized annotation efforts \cite{ehrmann_named_2024}. 

To address the gap of gold-standard NER for historical data in Brazilian Portuguese, we introduce MariNER (\textit{Mapeamento e Anotações de Registros Históricos para} NER, or Mapping and Annotations of Historical Records for NER). MariNER is the first dataset for Brazilian Portuguese NER focused on early-20th-century texts, and comprises five documents, including travel records and published historical articles, originally gathered for discourse analysis in a prior study \cite{omitted}. These documents, which primarily describe expeditions through the Brazilian North-east region, were digitized, normalized to standard Brazilian Portuguese, and manually annotated with four named entity categories: PERSON, LOCATION, ORGANIZATION, and DATE. The dataset comprises over 9,000 sentences, which may provide a robust foundation for developing NER models targeted towards historical texts.

With MariNER, we aim to support the development of tools for entity recognition in Portuguese historical texts, thereby advancing research in digital humanities and related fields. This work details the annotation process, including the challenges encountered and strategies to ensure high-quality annotations. We also present an analysis of the dataset’s characteristics, such as entity distributions and mention lengths, and compare them to existing Portuguese NER datasets. Finally, we evaluate the performance of state-of-the-art NER models on MariNER, establishing a benchmark for future research.

\section{Related Work}

The availability of Portuguese NER datasets remains limited compared to other languages, particularly for historical texts \cite{albuquerque_named_2023}. While existing resources cover modern domains \cite{santos_harem_nodate,villavicencio_lener-br_2018} and silver-standard automatically annotated collections \cite{higuchi_text_nodate}, there is a notable scarcity of manually annotated gold-standard datasets for historical Portuguese, as observed by \cite{santos_named_nodate}.

Recent efforts have attempted to address this gap. \cite{higuchi_text_nodate} developed a corpus of historical Brazilian Portuguese drawn from the \textit{Dicionário Histórico-Biográfico Brasileiro}, covering texts from Brazil's Old Republic (1930s). However, this corpus was annotated automatically, making it a silver-standard dataset. Although a subset was manually labelled for analysis, this portion is not publicly available.

More recently, \cite{santos_named_nodate} introduced a gold-standard corpus for Portuguese historical entity recognition based on the \textit{Parish Memories} Corpus, a collection of 18th-century Portuguese records. While this represents a significant contribution for Portuguese NER, it is limited to the European variant of Portuguese.

Currently, no gold-standard NER dataset exists for early 20th-century Brazilian Portuguese. This gap hinders the development of digital humanities tools for analyzing historical records from this pivotal era. Our work addresses this limitation by introducing a manually annotated gold-standard dataset supporting improved NER models and cultural-historical research.

\section{MariNER}

MariNER is a dataset for NER in Brazilian Portuguese within the historical domain, built from early 20th-century historical articles and travel records, mainly from Brazil's North-East. The documents were originally compiled for discourse analysis in \cite{omitted} and transcribed from non-digitized sources. This section details the data collection, entity definitions, annotation process, and dataset statistics.

\subsection{Data Collection}

The collected documents consist of various textual reports focusing on the countryside of Piauí, a state in Brazil's North-East region, but that also describe neighboring areas. These reports were authored by individuals affiliated with public and research institutions in Brazil during the early 20th century. The documents were originally compiled to analyze and better understand the narratives constructed around this territory at the time, such as common prejudices related to the region and the people that live in there. They were sourced as photographs of the original texts, from both the original institutions that commissioned their creation and current custodians of the texts. In \cite{omitted}, an initial annotation was provided for some entity types, which was then used to perform some quantitative analysis and build knowledge graphs to support the discourse analysis of the documents. In this work, we aimed to extend those annotations to be used in future works.
    
\subsection{Entity Types}

The entity types considered for this work are the ones that were considered most relevant for the nature of the analysis that can be performed on the underlying texts that compose the dataset. Initially, only entities of type PERSON and LOCATION were considered, as they were sufficient to perform a quantitative analysis of the underlying documents \cite{omitted}. Later, entities of types DATE and ORGANIZATION were also included, as studies indicated that these types of mentions might be relevant for future evaluations of the original documents and valuable resources for performing NER for other historical documents. %Table \ref{tab:repeated} shows the most common mentions that show up in the dataset.

The PERSON entity type covers explicit mentions to people, but not groups of people, occupations, or other collective denominations of people. Due to the documents' nature, first names tend to be more common, but mentions to full names and people with titles are also commonplace, such as ``\textit{Frei Cristovam Severim}'', or ``\textit{J. C. FLETCHER}.''

LOCATION covers mentions to places and geographical areas. The most common mentions are to cities, towns, or states, usually in the Brazilian north-eastern region. However, mentions to more specific places such as ``\textit{Porto Nacional}'' (National Port) or ``\textit{fazenda da Sussuarana}'' (\textit{Sussuarana} farm) are also common, with mentions to farms being frequent throughout the documents.

DATE refers to mentions to distinguishable dates, be it specific days, years, months, weekdays, or a general time in the year. As some of the documents are travel records in nature, mentions to years are commonplace.

Finally, ORGANIZATION mentions are related to clearly distinguishable organizations in the text. It means that, if the organization ``\textit{Instituto Oswaldo Cruz}'' (\textit{Oswaldo Cruz} Institute) is mentioned in full, but later, for example, referenced as ``\textit{o instituto}'' (the institute), only the first mention is labeled as en entity of this type.

\subsection{Text Pre-Processing}

As described in \cite{teixeira_sertoes_2024}, the original documents were digitized from photographs of their physical counterparts using Optical Character Recognition (OCR). Next, an annotator manually verified the OCR output to ensure accuracy. To eliminate spelling variations and inconsistencies, the texts were then normalized to standard Brazilian Portuguese, following a process similar to \cite{santos_named_nodate} in which dated spellings were converted to modern ones. To facilitate processing with automated NER systems, the documents were segmented into sentences using NLTK's\footnote{https://www.nltk.org/} Portuguese sentence tokenizer. By the end of the pre-processing stage, the original documents had been transformed into over 9,000 sentences in standardized Brazilian Portuguese.

\subsection{Annotation Guidelines}

Annotation guidelines were established to standardize the annotation process. The entity definitions were adapted from the ones used in the HAREM dataset \cite{santos_harem_nodate}, specifically from their annotation directives\footnote{https://www.linguateca.pt/aval\_conjunta/HAREM/directivas.html}, as our entity types align with those covered in their work. However, we identified certain mismatches between HAREM's definitions and the specific types of interest for our dataset. To address this, we made the following revisions to their guidelines for each entity type:

\begin{itemize}
    \item LOCATION: We do not consider the subtype \textit{REGIÃO} (Region) described in the guidelines, as we determine that this subtype would bring more complexity to the annotation task while not covering our annotation needs. Acronyms also had to be treated differently, with cities or towns named with their respective states' acronyms (e.g. \textit{Juazeiro (BA)} being labeled as a single entity, while cities or towns named with the states' full names (e.g. \textit{Juazeiro, Bahia}) being labeled as separate entities.

    \item DATE: For the DATE entity type, we mostly use HAREM's guidelines available for \textit{TEMPO} (TIME). However, we do not consider time spans in our annotations, as this would introduce unnecessary complexity for this entity type. It is for this reason that we decided to define this label as DATE rather than TIME. 
\end{itemize}

% With regards to the annotation process, we decide to perform the annotation in rounds, with each round being done by a single annotator and reviewed by the rest of the annotation team. The annotation process may be done through many rounds, but the final one would be a review stage done by the whole team.

\subsection{Manual Annotation Process}

The annotation process was carried out in three distinct phases by a team composed of three computer scientists and one historian-architect. The first round of annotations was primarily conducted for the analysis outlined in \cite{omitted}. During this phase, annotations were performed without explicit guidelines or a specialized NER tagging tool, and only PERSON and LOCATION NEs were considered. For the second round, annotations were conducted following the formal guidelines described earlier. This phase involved refining the existing PERSON and LOCATION annotations and introducing the DATE and ORGANIZATION entity types. Finally, a third round of annotation was performed, where the annotation team reviewed and evaluated ambiguous mentions obtained in the previous rounds. Unlike the first round, the second and third rounds utilized the LabelStudio\footnote{https://labelstud.io/} annotation tool.

\subsection{Dataset Statistics}

The final dataset comprises 9,649 labelled sentences, distributed into train, evaluation, and test sets with approximate proportions of 65\%, 15\%, and 20\%, respectively. Table \ref{tab:entities} shows a breakdown of the number of named entities in each split and the total counts across the full set. The dataset exhibits an imbalance in entity distribution, with LOCATION being the most frequent entity type, followed by PERSON. This distribution aligns with the nature of the original documents, which primarily focus on travel records and historical articles.

\begin{table}[htbp]
\centering
\caption{Number of entities per data split and entity type}
\label{tab:entities}
\begin{tabular}{lcccc}
\toprule
\textbf{Entity Type} & \textbf{Train} & \textbf{Eval} & \textbf{Test} & \textbf{Total} \\ \midrule
LOCATION                & 2697           & 773          & 958           & 4428           \\
ORGANIZATION          & 147            & 33           & 51            & 231            \\
DATE                 & 414            & 126          & 130           & 670            \\
PERSON               & 815            & 226          & 274           & 1315           \\ \midrule
\textbf{Total}       & \textbf{4073}  & \textbf{1158} & \textbf{1413} & \textbf{6644}  \\ \bottomrule
\end{tabular}
\end{table}

Table \ref{tab:repeated} presents the most frequent mentions for each entity type. The column \textit{Freq.} indicates the frequency of each mention within its respective entity type. These mentions show some partial insights into the content of the underlying documents. For instance, the most common location, \textit{Piauí}, reflects the focus of the documents on this northeastern Brazilian state. Similarly, the frequent dates align with the early 20th-century time frame of the travel records. %Moreover, MariNER exhibits unique mention length distributions compared to other Portuguese NER datasets. Figure \ref{fig:lengths} illustrates a comparison between MariNER, HAREM \cite{santos_harem_nodate}, and LeNER-BR \cite{villavicencio_lener-br_2018}, two datasets with similar entity types. Notably, organizations in MariNER tend to have slightly longer character lengths, while person mentions are generally shorter.

\begin{table}[H]
\centering
\caption{Frequency of most repeated mentions per entity type.}
\label{tab:repeated}
\begin{tabular}{l|lc}
\toprule
\textbf{Type} & \textbf{Mention} & \textbf{Freq. (\%)} \\ \midrule
LOCATION      & Piauí            & 5.31                   \\
              & Goiás            & 4.81                   \\
              & Bahia            & 3.77                   \\
              & Brasil           & 3.01                   \\
              & São Paulo        & 2.26                   \\ \midrule
PERSON        & Tamara           & 7.86                   \\
              & Brauni           & 4.11                   \\
              & Mário            & 3.82                   \\
              & Tânia            & 3.46                   \\
              & Pacheco          & 3.32                   \\ \midrule
DATE          & 1911             & 2.32                   \\
              & 1913             & 1.88                   \\
              & 1912             & 1.88                   \\
              & Agosto           & 1.59                   \\
              & 1914             & 1.45                   \\ \midrule
ORGANIZATION & Coral Popular    & 5.75                   \\
              & Parques Infantis & 5.31                   \\
              & Governo          & 4.87                   \\
              & Instituto Oswaldo Cruz & 3.10           \\
              & Parahyba-Hotel   & 3.10                   \\ \bottomrule
\end{tabular}
\end{table}

% \begin{figure}[htbp]
%     \centering
%     \includegraphics[width=0.8\linewidth]{figures/lengths.pdf}
%     \caption{Length distribution in characters of mentions per entity of different datasets. For HAREM and LeNER-BR, entities of type \textit{TEMPO} were compared for type DATE.}
%     \label{fig:lengths}
% \end{figure}

\section{NER Methods}

To evaluate the performance of automated named entity recognition systems for MariNER, we select 3 methods commonly seen in the literature, totaling 8 different models to perform NER for our dataset. In this section we go over these methods in more detail.

\subsection{LSTM-CRF Based}

\begin{figure}[htbp]
    \centering
    \includegraphics[width=0.8\linewidth]{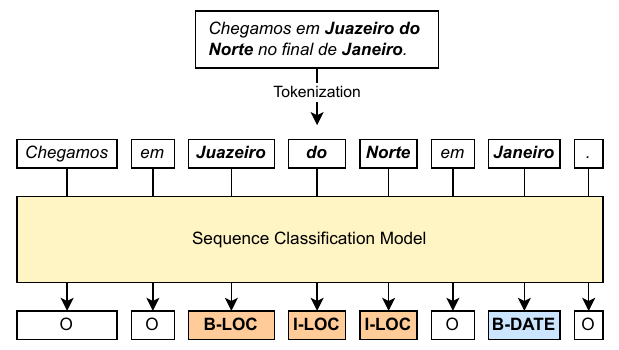}
    \caption{NER as a sequence classification task. The original sentence is split into tokens and each token is processed by the model, which outputs its class as one of the possible BIO tags. In the example, \textit{``Juazeiro do Norte''} is a location entity and \textit{``Janeiro''} is a date entity.}
    \label{fig:seq-class}
\end{figure}

In this method, the input sentences are pre-tokenized, and each token is annotated using the BIO format \cite{ramshaw_text_1995}. The task is framed as a sequence classification problem, where the goal is to assign the correct BIO label to every token in the sequence. Figure \ref{fig:seq-class} provides a visual representation of this task format.

The LSTM-CRF model is composed of two main parts: a Long Short-Term Memory (LSTM) \cite{hochreiter_long_1997} layer and a Conditional Random Fields (CRF) \cite{sutton_introduction_2012} classifier, a common statistical model for sequence labeling tasks such as NER. Word embeddings are obtained for each input token, which are then processed by the LSTM layer in order to learn patterns in the data. Finally, the outputs of the LSTM layer are fed into the CRF classifier to obtain the final label for the token.

\subsection{BERT Based}

The BERT family of models \cite{devlin_bert_2019} are pre-trained encoder language models that are popular to perform a great number of NLP tasks. Given an input text, these models are capable of generating contextualized embedding vectors for each to the text's tokens. By attaching classification modules to the end of the BERT models, it is possible to fine-tune these models and their embedding representations to perform NER as a \textbf{sequence classification task} \cite{santos_assessing_2019}, again using the BIO format.

For our experiments, we chose to use two distinct encoder models: BERTimbau \cite{souza_bert_2023}, a BERT model pre-trained on a Portuguese dataset; XML-RoBERTa \cite{conneau_unsupervised_2020}, a variation of the original BERT architecture pre-trained on a multilingual dataset that includes Portuguese. As the output of the encoder models is a contextualized word embedding, classification modules must be used to obtain a classification from those final representations. We used both a simple linear layer and a linear layer combined with a CRF classifier to perform the final classification of the tokens. 

\subsection{LLM Based}

Large Language Models (LLMs) have gained substantial popularity in the recent years for many NLP tasks. This popularity is due to the fact that these models show capabilities to perform tasks that they were not previously taught during their pre-training process \cite{zhao_survey_2025}. Not only that, but these models also seem to be capable to learn to perform said tasks better through a process named In Context Learning (ICL) \cite{zhao_survey_2025}. To achieve ICL, instructions and examples are passed to the model through an input prompt in natural language form. Similarly, the output of the model is also given in natural language. This configures most of the use cases of LLMs to perform arbitrary tasks as text-to-text pipelines.

\begin{figure}[htbp]
    \centering
    \includegraphics[width=0.8\linewidth]{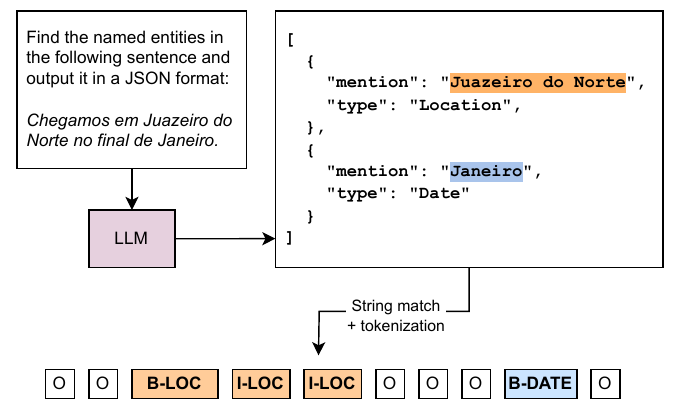}
    \caption{NER as a text-to-text task. A prompt is constructed requiring the LLM to extract the entities of the given sentence. The model outputs a JSON formatted output, which is parsed and matched with the original input to obtain its final BIO classification.}
    \label{fig:ner-llm}
\end{figure}

Various works attempt to convert the traditional sequence classification format of NER into a text-to-text task. Namely, \cite{santos_named_nodate} attempts to query the LLM to generate the original input sentence along with BIO tags for each token, while \cite{wang_gptner_2023} requires the model to only tag the desired entities with identification tags. \cite{xie_empirical_2023} queries the model to only generate the list of entities present in the text. Due to it not requiring that the model performs alignment internally to produce the final output, and due to it being simpler to parse and obtain the list of present entities, this last approach was chosen for our experiments. Figure \ref{fig:ner-llm} illustrates how we implemented this method. The LLM is queried to generate a list of entities and their types in JSON format. After parsing this structured output, string matching is performed to obtain the spans of the mentions in the original sentence. This can then be converted into the BIO format for easy comparison with the methods previously described.

\section{Experiments}

For our experiments, we train all of the previous models on MariNER's train set, evaluate to obtain the best configurations using the eval set and test the final performance using the test set. We also investigate the cross-dataset performance of the models when trained on a different source dataset and evaluated in MariNER. For this, we considered two datasets as the alternative source datasets:

\begin{itemize}
    \item HAREM \cite{santos_harem_nodate}, a popular general-domain Portuguese NER dataset that is annotated for entity types \textit{Person}, \textit{Date}, \textit{Location}, \textit{Organization} and \textit{Value}. This means that the set of entity types considered in MariNER is a subset of HAREM's;
    \item LeNER-BR \cite{villavicencio_lener-br_2018}, another popular Portuguese NER dataset of the legal domain. LeNER-BR is annotated for \textit{Person}, \textit{Location}, \textit{Organization}, \textit{Data}, \textit{Law} and \textit{Jurisprudence}, making its entity types also a superset of MariNER's. The Portuguese legal domain often uses formal and archaic language, which may have enough similarity with MariNER's language form to boost the NER model's performance. 
\end{itemize}

\subsection{Model Configurations}

For methods that treat NER as a sequence classification task, the \textit{Flair}\footnote{https://github.com/flairNLP/flair} framework was used, as it provides the models' architectures for easy usage. For the LSTM-CRF architecture, we follow the implementation and best model configurations for the NER task provided in \cite{santos_assessing_2019}, which uses \textit{Flair}'s \textit{FlairBBP} embeddings, combined with shallow 300-dimensional \textit{Glove} NILC embeddings\footnote{http://nilc.icmc.usp.br/nilc/index.php/repositorio-de-word-embeddings-do-nilc} with embedding stacking as the input for a Bi-Directional LSTM layer (BiLSTM-CRF+Flair). The model was trained for 150 epochs, with a batch size of 32 and learning rate of 0.1 with the Stochastic Gradient Descend optimizer. For the BERT based architectures, we used BERTimbau's\footnote{https://huggingface.co/neuralmind/bert-large-portuguese-cased} and XLM-RoBERTa's\footnote{https://huggingface.co/FacebookAI/xlm-roberta-large} large variations and fine-tuned them for 15 epochs and batch size of 8, with learning rate of $3\times10^{-5}$ using the AdamW optimizer.

For the LLM based approach, we evaluated three different language models: Llama3.1-8B-Instruct\footnote{https://huggingface.co/meta-llama/Meta-Llama-3-8B-Instruct} \cite{grattafiori_llama_2024}, Gemma2-9B-Instruct\footnote{https://huggingface.co/google/gemma-2-9b-it} \cite{team_gemma_2024} and Qwen2\footnote{https://huggingface.co/Qwen/Qwen2.5-7B-Instruct} \cite{yang_qwen2_2024}. These models were chosen because they feature diverse architectures and training processes while maintaining similar numbers of trainable parameters. For each model, ICL is used in zero-shot and $k$-shot approaches, with $k$ being the number of labelled examples shown to the LLM in the input prompt. We considered $k = 5, 10, 20$ for the experiments. The temperature $t$ for each model was kept at $t = 0$ as the list of extracted entities shouldn't vary across the model's text generation processes. We didn't use any quantization methods to instantiate the models.

For the methods that treat the task as sequence classification, experiments were ran three times and their average is reported. For text-to-text methods, experiments were ran once as the underlying LLM's temperature was kept at 0. We report the precision, recall and micro-F1 scores on MariNER's test set, which are calculated in the entity-level, similarly to how it's implemented in the CoNLL-2002 \cite{tjong_introduction_2002} NER evaluation scripts.

\subsection{Results}

Table \ref{tab:performance} presents the precision ($P$), recall ($R$), and micro-F1 ($F1$) scores of the evaluated methods and models across all entity types for MariNER's test set. The results indicate that XLM-RoBERTa ($XLM-R$) paired with a linear classifier is the best-performing model overall, achieving an average F1 score of $0.922$ across the entity types. While all BERT-based models exhibit similar performance levels, the BiLSTM-CRF+Flair ($BiLSTM$) model shows a more significant performance drop compared to its BERT-based counterparts, though this difference is still small at only $0.025$ for the average F1 score. Notably, however, all models struggle with the ORGANIZATION entity type, with the highest F1 score for this category being $0.640$, achieved by BERTimbau-CRF ($BERT-CRF$). This represents a significant drop of nearly $0.30$ compared to the best scores for other entity types, which indicates that recognizing organizations in historical texts is more challenging than other categories.

\begin{table}[htbp]
\caption{Performance of the different models for different entity types. Best results per column are highlighted in \textbf{bold}.}
\label{tab:performance}

\footnotesize
\centering
\begin{tabular}{@{}llcccccccc@{}}
\toprule
\multicolumn{2}{c|}{\multirow{2}{*}{Metric}}             & \multicolumn{8}{c}{Model}                                                                                                  \\ \cmidrule(l){3-10} 
% \multicolumn{2}{c}{}                                    & BiLSTM-CRF+Flair & BERTimbau      & BERTimbau-CRF  & XLM-RoBERTa    & XLM-RoBERTa-CRF & Qwen2          & Llama3.1 & Gemma2 \\ \midrule

\multicolumn{2}{c|}{}                                    & BiLSTM & BERT & BERT-CRF & XLM-R & XLM-CRF & Qwen2 & Llama3 & Gemma2 \\ \midrule

\multirow{3}{*}{LOC.}     & \multicolumn{1}{l|}{P}  & 0.887            & 0.927          & \textbf{0.934} & 0.931          & 0.927           & 0.510          & 0.491    & 0.704  \\
                              & \multicolumn{1}{l|}{R}  & 0.924            & 0.924          & 0.930          & \textbf{0.938} & 0.936           & 0.774          & 0.735    & 0.794  \\
                              & \multicolumn{1}{l|}{F1} & 0.905            & 0.926          & 0.932          & \textbf{0.935} & 0.932           & 0.615          & 0.589    & 0.746  \\ \midrule
\multirow{3}{*}{PERS.}       & \multicolumn{1}{l|}{P}  & 0.896            & 0.925          & \textbf{0.934} & 0.933          & 0.931           & 0.481          & 0.283    & 0.563  \\
                              & \multicolumn{1}{l|}{R}  & 0.896            & \textbf{0.928} & 0.925          & 0.927          & 0.912           & 0.724          & 0.713    & 0.765  \\
                              & \multicolumn{1}{l|}{F1} & 0.896            & 0.927          & 0.929          & \textbf{0.930} & 0.921           & 0.578          & 0.405    & 0.648  \\ \midrule
\multirow{3}{*}{DATE}         & \multicolumn{1}{l|}{P}  & 0.914            & 0.874          & 0.885          & \textbf{0.924} & 0.918           & 0.392          & 0.296    & 0.358  \\
                              & \multicolumn{1}{l|}{R}  & 0.972            & 0.958          & 0.953          & 0.972          & \textbf{0.977}  & 0.733          & 0.575    & 0.870  \\
                              & \multicolumn{1}{l|}{F1} & 0.942            & 0.914          & 0.918          & \textbf{0.947} & 0.947           & 0.511          & 0.391    & 0.507  \\ \midrule
\multirow{3}{*}{ORG.} & \multicolumn{1}{l|}{P}  & \textbf{0.696}   & 0.607          & 0.657          & 0.645          & 0.642           & 0.097          & 0.022    & 0.119  \\
                              & \multicolumn{1}{l|}{R}  & 0.571            & 0.625          & 0.625          & 0.577          & 0.637           & \textbf{0.682} & 0.591    & 0.636  \\
                              & \multicolumn{1}{l|}{F1} & 0.627            & 0.616          & \textbf{0.640} & 0.609          & 0.639           & 0.169          & 0.043    & 0.201  \\ \midrule
\multirow{3}{*}{Overall}      & \multicolumn{1}{l|}{P}  & 0.886            & 0.907          & 0.918          & \textbf{0.920} & 0.915           & 0.439          & 0.281    & 0.549  \\
                              & \multicolumn{1}{l|}{R}  & 0.909            & 0.916          & 0.919          & \textbf{0.925} & 0.923           & 0.756          & 0.709    & 0.791  \\
                              & \multicolumn{1}{l|}{F1}                      & 0.897            & 0.912          & 0.919          & \textbf{0.922} & 0.919           & 0.556          & 0.402    & 0.648  \\ \bottomrule
\end{tabular}

\end{table}

Large language models underperform compared to the smaller, specialized models, with Google's Gemma2 achieving the highest overall F1 score of $0.648$ among the LLMs. This aligns with existing findings that these models often struggle with specific tasks, such as named entity recognition, when compared to task-specific models \cite{wang_gptner_2023}. Additionally, the results suggest that LLMs tend to achieve higher recall but suffer from significantly lower precision. For instance, Qwen2 achieves a recall of $0.682$ for the ORGANIZATION entity type, the highest recall score among all models, but its precision drops to $0.097$, reducing its F1 score.

\subsection{K-Shot NER With LLMs}

\begin{figure}
    \centering
    \includegraphics[width=0.7\linewidth]{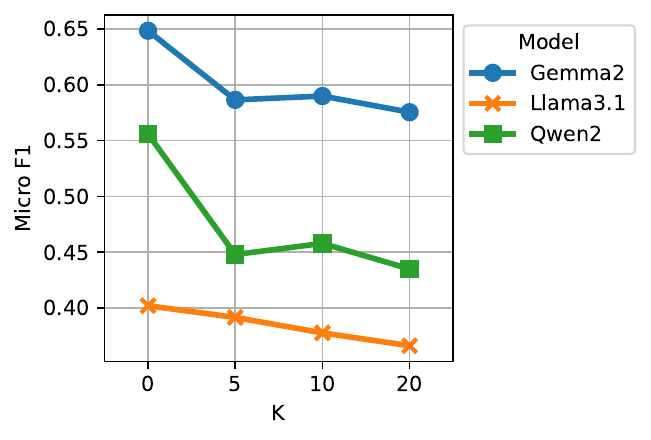}
    \caption{LLM performances for different $K$-shots.}
    \label{fig:shots}
\end{figure}

It is worth noting that the reported results for all LLMs are based on a zero-shot configuration, which consistently yielded the best performance across models. Figure \ref{fig:shots} illustrates the performance of LLMs across different $K$-shot configurations. Interestingly, adding few-shot examples to the input prompts appears to hinder performance rather than improve it. This may be attributed to the relatively smaller size of these locally-run LLMs, as the increased context size from additional examples could negatively affect their capacity to effectively process and extract entities.

\subsection{Cross Dataset Training}

Regarding cross-dataset performance, Table \ref{tab:cross-dataset} presents the overall micro-F1 scores for LSTM-CRF and BERT-based models. LLMs were excluded from this analysis, as the zero-shot configuration proved to be the most effective setup in prior experiments. The results demonstrate that the models can generalize information from the source dataset to the target dataset with reasonable performance. Notably, HAREM appears to be more informative for most models, as they consistently achieve higher performance on MariNER when trained on HAREM compared to LeNER-BR. This suggests that the linguistic proximity between early-20th-century texts and modern legal texts (as in LeNER-BR) may not provide significant benefits for model training. While the models were able to extract useful information from different source datasets, it is evident that domain-specific training data plays an important role in achieving high performances on MariNER, which indicates the importance of tailored datasets for historical NER tasks.

\begin{table}[htpb]
\centering
\caption{Overall performance of the different models for different source datasets. Best results per column per source set are highlighted in \textbf{bold}.}
\label{tab:cross-dataset}
\footnotesize
\begin{tabular}{l|l|lll}
\toprule
\multirow{2}{*}{Source} & \multicolumn{1}{c|}{\multirow{2}{*}{Model}} & \multicolumn{3}{c}{Overall} \\
 & \multicolumn{1}{c|}{} & \multicolumn{1}{c}{P} & \multicolumn{1}{c}{R} & \multicolumn{1}{c}{F1} \\ \midrule
\multirow{5}{*}{HAREM} & BiLSTM-CRF+Flair & 0.665 & 0.718 & 0.691 \\
 & BERTimbau & \textbf{0.679} & 0.785 & 0.728 \\
 & BERTimbau-CRF & 0.676 & \textbf{0.794} & \textbf{0.730} \\
 & XLM-RoBERTa & 0.670 & 0.759 & 0.712 \\
 & XLM-RoBERTa-CRF & 0.673 & 0.765 & 0.716 \\ \midrule
\multirow{5}{*}{LeNER-BR} & BiLSTM-CRF+Flair & 0.618 & 0.638 & 0.628 \\
 & BERTimbau & 0.647 & \textbf{0.712} & 0.677 \\
 & BERTimbau-CRF & \textbf{0.660} & 0.711 & \textbf{0.684} \\
 & XLM-RoBERTa & 0.602 & 0.650 & 0.625 \\
 & XLM-RoBERTa-CRF & 0.636 & 0.643 & 0.639 \\ \bottomrule
\end{tabular}
\end{table}

\section{Conclusion}

In this work, we presented MariNER, a gold-standard dataset for Named Entity Recognition in the Brazilian Portuguese historical domain. Comprising five distinct documents, ranging from historical articles to travel records, the dataset focuses on early 20th-century expeditions, mostly through Brazil's North-East region. With over 9,000 annotated sentences, MariNER is comparable in size to existing Portuguese NER datasets, making it an important resource for research and practical applications.

Experiments with state-of-the-art NER models showed that XLM-RoBERTa achieves the best performance on MariNER. We also found that LLMs underperformed when prompted to extract named entities from the dataset, with the best results obtained using Gemma2 in a zero-shot configuration, but still with an almost 30\% gap in micro-F1 performance when compared to fine-tuned models. Cross-dataset training experiments revealed that while models can learn from other datasets, they perform almost 20\% better when trained on MariNER's training data, which highlights the importance of domain-specific annotations for historical texts.

Future work includes expanding entity definitions to include sub-types (e.g., differentiate between cities, states, and countries for the LOCATION type) to provide finer-grained annotations for more specific extraction tasks. Another possibility would be increasing the number and diversity of source documents, by including travel records from expeditions to other Brazilian regions for example, in order to better capture linguistic and contextual variations in early 20th-century Brazilian Portuguese.

\bibliographystyle{splncs04}
\bibliography{references}

\begin{thebibliography}{10}
\providecommand{\url}[1]{\texttt{#1}}
\providecommand{\urlprefix}{URL }
\providecommand{\doi}[1]{https://doi.org/#1}

\bibitem{albuquerque_named_2023}
Albuquerque, H.O., Souza, E., Gomes, C., de~C.~Pinto, M.H., Filho, R.P.S., Costa, R., de~M.~Lopes, V.T., da~Silva, N.F.F., de~Carvalho, A.C.P.L.F., Oliveira, A.L.I.: Named {Entity} {Recognition}: a {Survey} for the {Portuguese} {Language}. Procesamiento del Lenguaje Natural pp. 171--185 (Mar 2023). \doi{10.26342/2023-70-14}, \url{https://doi.org/10.26342/2023-70-14}

\bibitem{conneau_unsupervised_2020}
Conneau, A., Khandelwal, K., Goyal, N., Chaudhary, V., Wenzek, G., Guzm{\'a}n, F., Grave, E., Ott, M., Zettlemoyer, L., Stoyanov, V.: Unsupervised {{Cross-lingual Representation Learning}} at {{Scale}}. In: Proceedings of the 58th {{Annual Meeting}} of the {{Association}} for {{Computational Linguistics}}. pp. 8440--8451. Association for Computational Linguistics, Online (2020). \doi{10.18653/v1/2020.acl-main.747}

\bibitem{devlin_bert_2019}
Devlin, J., Chang, M.W., Lee, K., Toutanova, K.: {{BERT}}: {{Pre-training}} of {{Deep Bidirectional Transformers}} for {{Language Understanding}} (May 2019). \doi{10.48550/arXiv.1810.04805}

\bibitem{ehrmann_named_2024}
Ehrmann, M., Hamdi, A., Pontes, E.L., Romanello, M., Doucet, A.: Named {{Entity Recognition}} and {{Classification}} in {{Historical Documents}}: {{A Survey}}. ACM Computing Surveys  \textbf{56}(2),  1--47 (Feb 2024). \doi{10.1145/3604931}

\bibitem{grattafiori_llama_2024}
Grattafiori, A., Dubey, A., Jauhri, A., Pandey, A., Kadian, A., Al-Dahle, A., Letman, A., Mathur, A., Schelten, A., Vaughan, A., Yang, A., Fan, A., et~al.: The {{Llama}} 3 {{Herd}} of {{Models}} (Nov 2024). \doi{10.48550/arXiv.2407.21783}

\bibitem{higuchi_text_nodate}
Higuchi, S., Freitas, C., Cuconato, B., Rademaker, A.: Text {Mining} for {History}: ﬁrst steps on building a large dataset. In: Proceedings of the Eleventh International Conference on Language Resources and Evaluation ({LREC} 2018) (May 2018)

\bibitem{hochreiter_long_1997}
Hochreiter, S., Schmidhuber, J.: Long short-term memory. Neural Comput.  \textbf{9}(8),  1735–1780 (Nov 1997). \doi{10.1162/neco.1997.9.8.1735}, \url{https://doi.org/10.1162/neco.1997.9.8.1735}

\bibitem{li_survey_2022}
Li, J., Sun, A., Han, J., Li, C.: A {{Survey}} on {{Deep Learning}} for {{Named Entity Recognition}}. IEEE Transactions on Knowledge and Data Engineering  \textbf{34}(1),  50--70 (Jan 2022). \doi{10.1109/TKDE.2020.2981314}

\bibitem{villavicencio_lener-br_2018}
Luz De~Araujo, P.H., De~Campos, T.E., De~Oliveira, R.R.R., Stauffer, M., Couto, S., Bermejo, P.: {LeNER}-{Br}: {A} {Dataset} for {Named} {Entity} {Recognition} in {Brazilian} {Legal} {Text}. In: Villavicencio, A., Moreira, V., Abad, A., Caseli, H., Gamallo, P., Ramisch, C., Gonçalo~Oliveira, H., Paetzold, G.H. (eds.) Computational {Processing} of the {Portuguese} {Language}, vol. 11122, pp. 313--323. Springer International Publishing, Cham (2018). \doi{10.1007/978-3-319-99722-3_32}, \url{https://link.springer.com/10.1007/978-3-319-99722-3\_32}, series Title: Lecture Notes in Computer Science

\bibitem{oliveira_semclinbr_2022}
Oliveira, L.E.S.E., Peters, A.C., Da~Silva, A.M.P., Gebeluca, C.P., Gumiel, Y.B., Cintho, L.M.M., Carvalho, D.R., Al~Hasan, S., Moro, C.M.C.: {{SemClinBr}} - a multi-institutional and multi-specialty semantically annotated corpus for {{Portuguese}} clinical {{NLP}} tasks. Journal of Biomedical Semantics  \textbf{13}(1), ~13 (Dec 2022). \doi{10.1186/s13326-022-00269-1}

\bibitem{omitted}
Omitted: The author of this thesis was omitted to avoid misinterpretations during the double-anonymized reviewing process. Ph.D. thesis, University, Brazil (2024)

\bibitem{ramshaw_text_1995}
Ramshaw, L.A., Marcus, M.P.: Text {{Chunking}} using {{Transformation-Based Learning}} (May 1995). \doi{10.48550/arXiv.cmp-lg/9505040}

\bibitem{santos_harem_nodate}
Santos, D., Seco, N., Cardoso, N., Vilela, R.: {HAREM}: An advanced {NER} evaluation contest for {P}ortuguese. In: Calzolari, N., Choukri, K., Gangemi, A., Maegaard, B., Mariani, J., Odijk, J., Tapias, D. (eds.) Proceedings of the Fifth International Conference on Language Resources and Evaluation ({LREC}{'}06). European Language Resources Association (ELRA), Genoa, Italy (May 2006), \url{https://aclanthology.org/L06-1027/}

\bibitem{santos_named_nodate}
Santos, J., Cameron, H.F., Olival, F., Farrica, F., Vieira, R.: Named entity recognition specialised for {P}ortuguese 18th-century history research. In: Gamallo, P., Claro, D., Teixeira, A., Real, L., Garcia, M., Oliveira, H.G., Amaro, R. (eds.) Proceedings of the 16th International Conference on Computational Processing of Portuguese - Vol. 1. pp. 117--126. Association for Computational Lingustics, Santiago de Compostela, Galicia/Spain (Mar 2024), \url{https://aclanthology.org/2024.propor-1.12/}

\bibitem{santos_assessing_2019}
Santos, J., Consoli, B., Dos~Santos, C., Terra, J., Collonini, S., Vieira, R.: Assessing the {{Impact}} of {{Contextual Embeddings}} for {{Portuguese Named Entity Recognition}}. In: 2019 8th {{Brazilian Conference}} on {{Intelligent Systems}} ({{BRACIS}}). pp. 437--442. IEEE, Salvador, Brazil (Oct 2019). \doi{10.1109/BRACIS.2019.00083}

\bibitem{souza_bert_2023}
Souza, F., Nogueira, R., Lotufo, R.: {{BERT}} models for {{Brazilian Portuguese}}: {{Pretraining}}, evaluation and tokenization analysis. Applied Soft Computing  \textbf{149},  110901 (Dec 2023). \doi{10.1016/j.asoc.2023.110901}

\bibitem{sutton_introduction_2012}
Sutton, C.: An {{Introduction}} to {{Conditional Random Fields}}. Foundations and Trends{\textregistered} in Machine Learning  \textbf{4}(4),  267--373 (2012). \doi{10.1561/2200000013}

\bibitem{tamper_visualizing_2023}
Tamper, M., Leskinen, P., Hyv{\"o}nen, E.: Visualizing and {{Analyzing Networks}} of {{Named Entities}} in {{Biographical Dictionaries}} for {{Digital Humanities Research}}. In: Gelbukh, A. (ed.) Computational {{Linguistics}} and {{Intelligent Text Processing}}, vol. 13451, pp. 199--214. Springer Nature Switzerland, Cham (2023). \doi{10.1007/978-3-031-24337-0_15}

\bibitem{team_gemma_2024}
Team, G., Riviere, M., Pathak, S., Sessa, P.G., Hardin, C., Bhupatiraju, S., Hussenot, L., Mesnard, T., Shahriari, B., Ram{\'e}, A., Ferret, J., Liu, P., Tafti, P., et~al.: Gemma 2: {{Improving Open Language Models}} at a {{Practical Size}} (Oct 2024). \doi{10.48550/arXiv.2408.00118}

\bibitem{teixeira_sertoes_2024}
Teixeira, M.L.G.: Sertões do Piauhy: a construção das narrativas pela literatura de viagem (1912-1938). Tese (doutorado em teoria e história da arquitetura e do urbanismo), Instituto de Arquitetura e Urbanismo, Universidade de São Paulo, São Carlos (2024), [no prelo]

\bibitem{tjong_introduction_2002}
Tjong Kim~Sang, E.F.: Introduction to the {C}o{NLL}-2002 shared task: Language-independent named entity recognition. In: {COLING}-02: The 6th Conference on Natural Language Learning 2002 ({C}o{NLL}-2002) (2002), \url{https://aclanthology.org/W02-2024/}

\bibitem{wang_gptner_2023}
Wang, S., Sun, X., Li, X., Ouyang, R., Wu, F., Zhang, T., Li, J., Wang, G.: {{GPT-NER}}: {{Named Entity Recognition}} via {{Large Language Models}} (Oct 2023). \doi{10.48550/arXiv.2304.10428}

\bibitem{xie_empirical_2023}
Xie, T., Li, Q., Zhang, J., Zhang, Y., Liu, Z., Wang, H.: Empirical {{Study}} of {{Zero-Shot NER}} with {{ChatGPT}} (Oct 2023). \doi{10.48550/arXiv.2310.10035}

\bibitem{yang_qwen2_2024}
Yang, A., Yang, B., Hui, B., Zheng, B., Yu, B., Zhou, C., Li, C., Li, C., Liu, D., Huang, F., Dong, G., Wei, H., Lin, H., Tang, J., Wang, J., Yang, J., Tu, J., Zhang, J., Ma, J., Yang, J., Xu, J., Zhou, J., Bai, J., He, J., Lin, J., Dang, K., Lu, K., Chen, K., Yang, K., Li, M., Xue, M., Ni, N., Zhang, P., Wang, P., Peng, R., Men, R., Gao, R., Lin, R., Wang, S., Bai, S., Tan, S., Zhu, T., Li, T., Liu, T., Ge, W., Deng, X., Zhou, X., Ren, X., Zhang, X., Wei, X., Ren, X., Liu, X., Fan, Y., Yao, Y., Zhang, Y., Wan, Y., Chu, Y., Liu, Y., Cui, Z., Zhang, Z., Guo, Z., Fan, Z.: Qwen2 {{Technical Report}} (Sep 2024). \doi{10.48550/arXiv.2407.10671}

\bibitem{zhao_survey_2025}
Zhao, W.X., Zhou, K., Li, J., Tang, T., Wang, X., Hou, Y., Min, Y., Zhang, B., Zhang, J., Dong, Z., Du, Y., Yang, C., Chen, Y., Chen, Z., Jiang, J., Ren, R., Li, Y., Tang, X., Liu, Z., Liu, P., Nie, J.Y., Wen, J.R.: A {{Survey}} of {{Large Language Models}} (Mar 2025). \doi{10.48550/arXiv.2303.18223}

\end{thebibliography}




% LNCS Homepage, \url{http://www.springer.com/lncs}, last accessed 2023/10/25
% \end{thebibliography}
\end{document}